\documentclass{article}

\usepackage[preprint]{neurips_2021}

\usepackage[utf8]{inputenc} %
\usepackage[T1]{fontenc}    %
\usepackage{url}            %
\usepackage{booktabs}       %
\usepackage{amsfonts}       %
\usepackage{nicefrac}       %
\usepackage{microtype}      %
\usepackage{xcolor}         %
\usepackage{graphicx}
\usepackage{bm}
\usepackage{multirow}
\usepackage{caption}
\usepackage{subfig}

\usepackage[pagebackref=true,breaklinks=true,letterpaper=true,colorlinks,citecolor=blue,linkcolor=blue,bookmarks=false]{hyperref}
\newcommand{\bsigma}{\boldsymbol{\sigma}}

\usepackage{xcolor}
\usepackage{array}
\usepackage{xspace}

\def\signed #1{{\leavevmode\unskip\nobreak\hfil\penalty50\hskip2em
  \hbox{}\nobreak\hfil -- #1%
  \parfillskip=0pt \finalhyphendemerits=0 \endgraf}}
\newsavebox\mybox

\definecolor{ImproveGreen}{rgb}{0.2235, 0.7094, 0.2902}

\usepackage{color}
\definecolor{crimson}{rgb}{0.86, 0.08, 0.24}
\definecolor{gray}{rgb}{0.5,0.5,0.5}
\definecolor{green}{rgb}{0, 0.4, 0}
\definecolor{orange}{rgb}{1, 0.5, 0}
\definecolor{mahogany}{rgb}{0.75, 0.25, 0.0}
\definecolor{purple}{rgb}{0.6, 0, 0.6}
\definecolor{darkgreen}{rgb}{0, 0.4, 0}
\definecolor{frenchblue}{rgb}{0.0, 0.45, 0.73}
\definecolor{red}{rgb}{1,0,0}
\definecolor{yellow}{rgb}{1,1,0}
\definecolor{magenta}{rgb}{1,0,1}
\definecolor{pink}{rgb}{1,0.412,0.706}

\newcommand{\mycomment}[1]{}
\newcommand{\comment}[1]{}

\makeatletter
\DeclareRobustCommand\onedot{\futurelet\@let@token\@onedot}
\def\@onedot{\ifx\@let@token.\else.\null\fi\xspace}

\makeatother

\newlength\paramargin
\newlength\figmargin
\newlength\subfigmargin
\newlength\secmargin
\newlength\subsecmargin
\newlength\tabmargin
\newlength\eqmargin

\newcommand{\subsecref}[1]{Section~\ref{subsec:#1}}
\newcommand{\figref}[1]{Figure~\ref{fig:#1}}

\long\def\ignorethis#1{}
\definecolor{crimson}{rgb}{0.86, 0.08, 0.24}
\definecolor{green}{rgb}{0, 0.5, 0.25}
\definecolor{purple}{rgb}{0.75, 0, 1}
\definecolor{orange}{rgb}{1, 0.5, 0.25}
\definecolor{yellow}{rgb}{1, 1, 0}
\definecolor{new_blue}{rgb}{0, 0.5, 1}

\title{StyleGAN of All Trades: Image Manipulation with Only Pretrained StyleGAN }

\author{Min Jin Chong$^{1}$\\
{\tt\small mchong6@illinois.edu}
\and
Hsin-Ying Lee$^{2}$\\
{\tt\small hlee5@snap.com}
\and
David Forsyth$^{1}$\\ 
{\tt\small daf@illinois.edu}\\
\and
$^1${University of Illinois at Urbana-Champaign} \quad
$^2${Snap Inc.} 
}

\begin{document}

\maketitle

\begin{abstract}
\vspace{\secmargin}
Recently, StyleGAN has enabled various image manipulation and editing tasks thanks to the high-quality generation and the disentangled latent space.
However, additional architectures or task-specific training paradigms are usually required for different tasks.
In this work, we take a deeper look at the spatial properties of StyleGAN.
We show that with a pretrained  StyleGAN along with some operations, without any additional architecture, we can perform comparably to the state-of-the-art methods on various tasks, including image blending, panorama generation, generation from a single image, controllable and local multimodal image to image translation, and attributes transfer.
The proposed method is simple, effective, efficient, and applicable to any existing pretrained StyleGAN model.
\end{abstract}
\begin{figure}[ht!]
    \centering
    \includegraphics[width=1\linewidth]{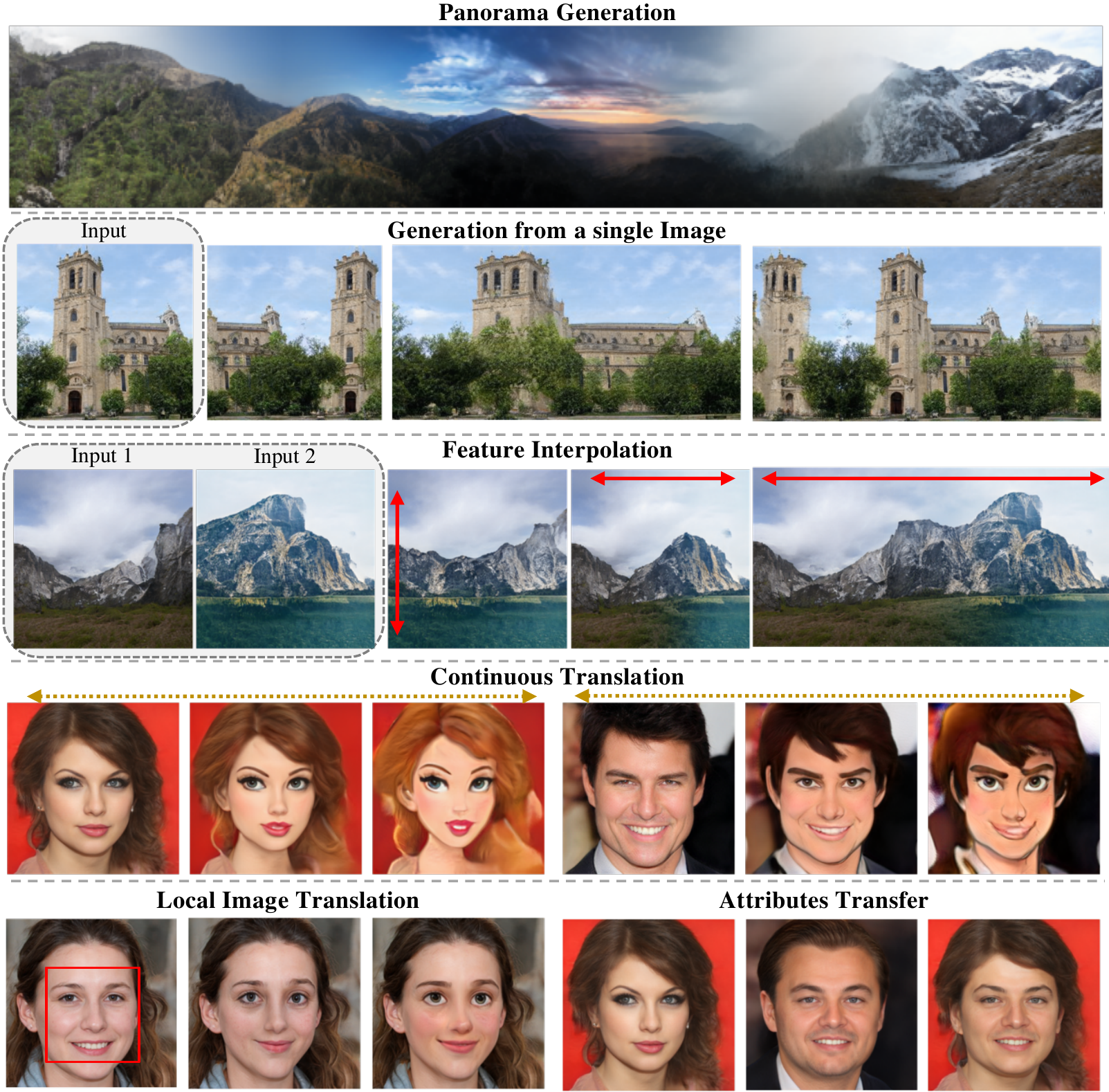}
    \vspace{-0mm}
    \caption{
    \textbf{Vanilla StyleGAN is all you need}.
    We use vanilla StyleGAN2 without any additional architectures to achieve various different tasks.
    }
    \label{fig:teaser}
    \vspace{\figmargin}
\end{figure}

\vspace{-5mm}
\section{Introduction}
\vspace{\secmargin}
Generative Adversarial Networks (GANs)~\cite{goodfellow2014generative} have made great progress in the field of image and video synthesis. 
Among all GANs models, recent StyleGAN~\cite{karras2019style} and StyleGAN2~\cite{karras2020analyzing} have further pushed forward the quality of generated images.
The most distinguishing characteristic of StyleGAN is the design of intermediate latent space that enables disentanglement of different attributes at different semantic levels.
This has attracted attention in trying to demystify the latent space and achieve simple image manipulations~\cite{shen2020interfacegan, shen2020interpreting, collins2020editing, abdal2019image2stylegan, wu2020stylespace}.

With its disentanglement property, StyleGAN has unleashed numerous image editing and manipulation tasks.
We can improve the controllability of the generation process via exploiting the latent space by augmenting and regularizing the latent space~\cite{chen2020free, alharbi2020disentangled, shoshan2021gan}, and by inverting images back to the latent space~\cite{abdal2019image2stylegan,bau2020rewriting,bau2020semantic}.
Furthermore, various conventional conditional image generation tasks can be achieved with the help of the inversion techniques. 
For example, image-to-image translation can be done by injecting encoded features to StyleGANs~\cite{richardson2020encoding,kwong2021unsupervised}, and image inpainting and outpainting can realized by locating the appropriate codes in the latent space~\cite{abdal2020image2stylegan++,cheng2021out, lin2021infinity}.
However, most methods either are designed in a task-specific manner or require additional architectures.

In this work, we demonstrate that a vanilla StyleGAN is sufficient to host a variety of different tasks, as shown in \figref{teaser}.
By exploiting the spatial properties of the intermediate layers along with some simple operations, we can, without any additional training, perform feature interpolation, panorama generation, and generation from a single image. 
With fine-tuning, we can achieve image-to-image translation which leads to various applications including continuous translation, local image translation, and attributes transfer.
Qualitative and quantitative comparisons show that the proposed method performs comparably to current state-of-the-art methods without any additional architecture.
All codes and models can be found at \url{https://github.com/mchong6/SOAT}.

\vspace{\secmargin}
\section{Related Work}
\vspace{\secmargin}
\paragraph{Image editing with StyleGAN}

The style-based generators of StyleGAN~\cite{karras2019style,karras2020analyzing} provide an intermediate latent space $\mathcal{W}+$ that has been shown to be semantically disentangled.
This property facilitates various image editing applications via the manipulation of the $\mathcal{W}+$ space.
In the presence of labels in the form of binary attributes or segmentations, vector directions in the $\mathcal{W}+$ space can be discovered for semantic edits \cite{shen2020interfacegan,wu2020stylespace}. 
In an unsupervised setting, EIS~\cite{collins2020editing} analyzes the style space of a large number of images to build a catalog that isolates and bonds specific parts of the style code to specific facial parts. 
However, as the latent space of StyleGAN is one-dimensional, these methods usually have limited control over spatial editing of images.
On the other hand, optimization-based methods provide better control over spatial editing.
Bau et al.~\cite{bau2020rewriting} allow users to interactively rewrite the rules of a generative model by manipulating the layers of a GAN as a linear associate memory. %
However, it requires optimization of the model weights and cannot work on the feature layers. 
To enable intuitive spatial editing, Suzuki et al.~\cite{suzuki2018spatially} perform collaging (cut and paste) in the intermediate spatial feature space of GANs, yielding realistic blending of images. However, due to the nature of collaging, the method are highly dependent on the pose and structure of the images and do not generate realistic results in many scenarios.

\vspace{\paramargin}
\paragraph{Panorama generation}
Panorama generation aims to generate a sequence of continuous images in an unconditional setting~\cite{lin2021infinity,lin2019coco, skorokhodov2021aligning} or conditioning on given images~\cite{cheng2021out}. 
These methods perform generation conditioning on a coordinate system. Arbitrary-lengthed panorama generation is then done by continually sampling along the coordinate grid.%

\vspace{\paramargin}
\paragraph{Generation from a single image}
SinGAN~\cite{rottshaham2019singan} recently proposes to learn the distribution of patches within a single image.
The learned distribution enable the generation of diverse samples that follows the patch distribution of the original image.
However, scalability is a major downside for SinGAN as every new image requires an individual SinGAN model, which is both time and computationally intensive.
On the contrary, the proposed method can achieve similar effect by manipulating the feature space of a pretrained StyleGAN.

\vspace{\paramargin}
\paragraph{Image to image translation (I2I)}
I2I aims to learn the mapping among different domains.
Most I2I methods~\cite{isola2017image,CycleGAN2017,huang2018multimodal,lee2018diverse} formulate the the mapping via learning a conditional distribution.
However, this formulation is sensitive to and heavily dependent on the input distribution, which often leads to unstable training and unsatisfactory inference results.
To leverage the unconditional distribution of both source and target domains, recently, 
Toonify~\cite{pinkney2020resolution} proposes to finetune a pretrained StyleGAN and perform weight swapping between the pretrained and finetuned model to allow high-quality I2I translation. 
Finetuning from a pretrained model allows the semantics learned from the original dataset to be well preserved. Less data is also needed for training due to transfer learning.
However, Toonify has limited controls and fails to achieve editing such as local translation and continouous translation.

\vspace{-5mm}
\vspace{\secmargin}
\section{Image Manipulation with StyleGAN}\label{sec:methods}
\vspace{0mm}
\vspace{\secmargin}
We introduce some common operations and their applications using StyleGAN. 
For the rest of the paper, let $f_i \in \mathcal{R}^{B\times C\times H\times W}$ represents intermediate features of the the $i$-th layer in the StyleGAN. 

\vspace{\subsecmargin}
\subsection{Setup}
\vspace{\subsecmargin}
All images generated are of $256 \times 256$ resolution.
For faces, we use the pretrained FFHQ model by rosinality~\cite{rosinality}; for churches, the pretrained model on the LSUN-Churches dataset~\cite{yu2015lsun} by Karras et al.~\cite{karras2020analyzing}; for landscapes and towers, we trained a StyleGAN2 model on LHQ~\cite{skorokhodov2021aligning} and LSUN-Towers~\cite{yu2015lsun} using standard hyperparameters. For face2disney and face2anime tasks, we fine-tune the FFHQ model on the Disney~\cite{pinkney2020resolution} and Danbooru2018~\cite{danbooru2018} dataset respectively. 

For quantitative evaluations, we perform user study and FID computations. All FID computations are implemented using the FID$_\infty$ by Chong et al.~\cite{chong2020effectively} which debiases the computation of FID. For our user study, given a pair of images, users are asked to choose the one that is more realistic and more relevant to the  task.
We ask each user to compare $25$ pairs of images from different methods and collect results from a total of $40$ subjects.
\begin{figure}[t]
    \centering
    \includegraphics[width=0.95\linewidth]{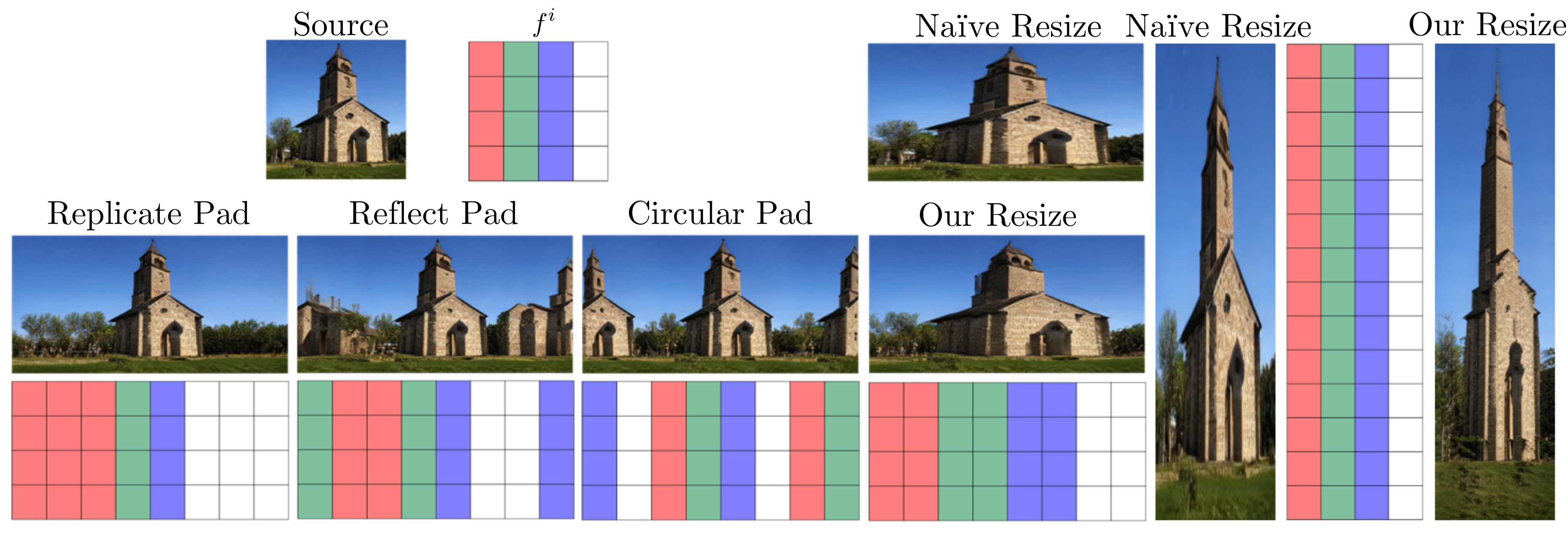}
    \vspace{-0mm}
    \caption{
    \textbf{Spatial Operation.}
    Performing simple spatial operations such as resizing and padding on StyleGAN's intermediate feature layers results in intuitive and realistic manipulations.
    }
    \label{fig:spatial_ops}
    \vspace{\figmargin}
\end{figure}

\vspace{\subsecmargin}
\vspace{-1mm}
\subsection{Simple spatial operations}
\vspace{\subsecmargin}
\vspace{-0mm}
Since StyleGAN is fully convolutional, we can adjust the spatial dimensions of $f_i$ to cause a corresponding spatial change in the output image. We experiment with simple spatial operations such as padding and resizing and show that we are able to achieve pleasing and intuitive results.

We apply all spatial operations on $f_2$. 
First, we perform padding operation that expands an input tensor by appending additional values to the borders of the tensor. In Fig.~\ref{fig:spatial_ops}, we explore several variants of paddings and investigate the results they have on the generated image. 
Replicate padding pads the tensor to its desired size by its boundary value. Fig.~\ref{fig:spatial_ops} shows that the background is extended d by replicating the bushes and trees.
Reflection padding reflects from the border, and
Circular padding wraps the tensor around, creating copies of the same tensor, as shown in Fig.~\ref{fig:spatial_ops}.
Then we introduce the resizing operation that performs resizing in the feature space.
Compared to naive resizing that causes artifacts such as blurred textures, resizing in the feature space maintains realistic texture.

\begin{figure}[t]
    \centering
    \includegraphics[width=1\linewidth]{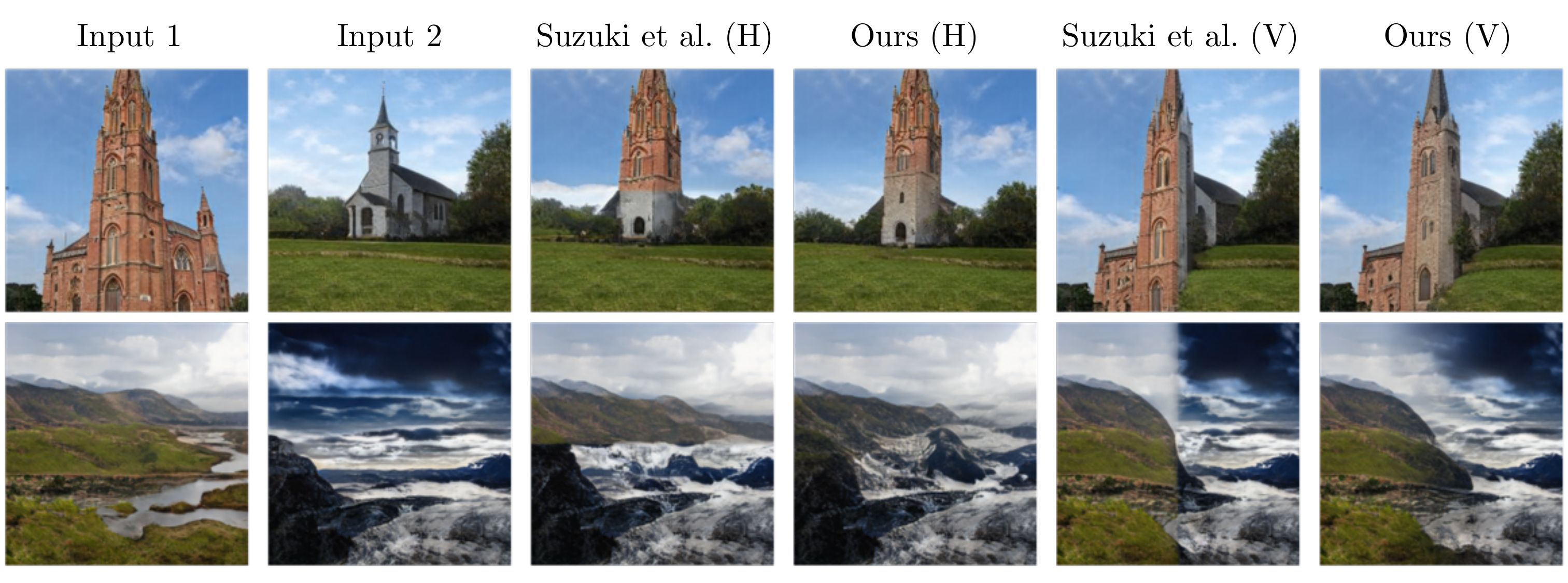}
    \caption{
    \textbf{Feature Interpolation.}
    We compare our feature interpolation with the feature collaging in Suzuki et al.~\cite{suzuki2018spatially}. H and V represent horizontal and vertical blending respectively. Our feature interpolation is able to blend and transition the two images seamlessly while there are obvious blending artifacts in Suzuki et al.
    }
    \label{fig:feature_blend}
    \vspace{\figmargin}
\end{figure}

\vspace{\subsecmargin}
\subsection{Feature interpolation}
\vspace{\subsecmargin}

Suzuki et al.~\cite{suzuki2018spatially} show that collaging (copy and pasting) features in the intermediate layers of StyleGAN allows the images to be blended seamlessly. However, this collaging does not work well when the images to be blended are too different.
Instead of collaging, we show that interpolating the features leads to smooth transitions between two images even if they are largely different. 

At each StyleGAN layer, we generate $f^A_i$ and $f^B_i$ separately using different latent noise. We then blend them smoothly with $f_i = (1-\alpha) f^A_i + \alpha f^B_i$,
where $\alpha \in [0,1]^{B,C,H,W}$ is a mask that blends the two features decided by different ways of blending, e.g. if for horizontal blending, the mask will get larger from left to right. $f_i$ is then passed on to the next convolution layer where the same blending will occur again. Note that we do not have to perform this blending at every single layer. We later show that strategic choices of where to blend can impact the results we get.

In most experiments, we set $\alpha$ linearly scaled using linspace which allows a smooth interpolation between the two features. The scale depends on the tasks. For landscapes, the two images are normally structurally different, and thus, benefit from a longer and slower scale that allows a smooth transition. This is evident in Fig.~\ref{fig:feature_blend}, where we compare feature interpolation with feature collaging in Suzuki et al. which fails to perform  smooth transition. 
We also perform a user study to let users select which interpoloated images look more realistic. As shown in Table~\ref{tab:user_study}, $87.6\%$ of users prefer our method  against Suzuki et al. .

\begin{figure}[t]
    \centering
    \includegraphics[width=1\linewidth]{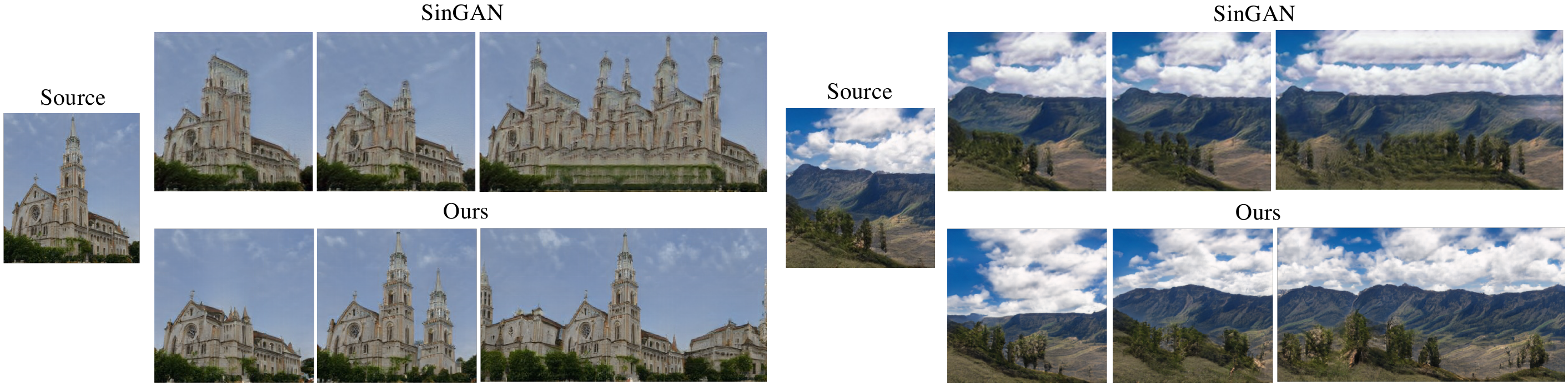}
    \vspace{-0mm}
    \caption{
    \textbf{Generation from a single image.}
    We compare single image generation with SinGAN. For our method, we perform feature interpolation to collage image structures or spatial paddings to widen the image. Our images are significantly more diverse and realistic. SinGAN fails to vary the church structures in meaningful way and generates unrealistic clouds and landscapes.
    }
    \label{fig:compare_singan}
    \vspace{\figmargin}
\end{figure}

\vspace{\subsecmargin}
\subsection{Generation from a single image}
\vspace{\subsecmargin}
In addition to feature interpolation between different images, we can apply interpolation within a single image. In some feature layers, we select relevant patches and replicate it spatially by blending it with other regions. Specifically, with a shift operator $\textrm{Shift}(\cdot)$ that translates the mask in a given direction:
\vspace{-1mm}
\begin{equation}
    f_i = \textrm{Shift}\big((1-\alpha)\big) f_i + \textrm{Shift}(\alpha f_i),
    \vspace{-0mm}
\end{equation}
In combination with simple spatial operations,  we can generate diverse images from a single image that has consistent patch distributions and structure. This is a similar task to SinGAN~\cite{rottshaham2019singan} with the exception that SinGAN involves sampling while we require manual choosing of patches for feature interpolation. Different from SinGAN that each image requires an individual model, our method uses the same StyleGAN with different latent codes.

We qualitatively and quantitatively compare the capability to generate from a single image of SinGAN and the proposed method. In Fig.~\ref{fig:compare_singan}, we perform comparisons on the LSUN-Churches and LHQ datasets. Our method generates realistic structures borrowed from different parts of the image and blends them into a coherent image. While SinGAN has more flexibility and is able to generate more arbitrary structures, in practice, the results are less realistic, especially in the case of  image extension. Notice in landscape extension, SinGAN is not able to correctly capture the structure of clouds, leading to unrealistic samples. Comparatively, the extension of our method based on reflection padding generates realistic textures that are structurally sound. For user study, we compare with SinGAN for image extension, with our method using spatial reflect padding at $f_2$. From Table~\ref{tab:user_study}, over 80\% of the users prefer our method.

\begin{figure}[t]
    \centering
    \includegraphics[width=1\linewidth]{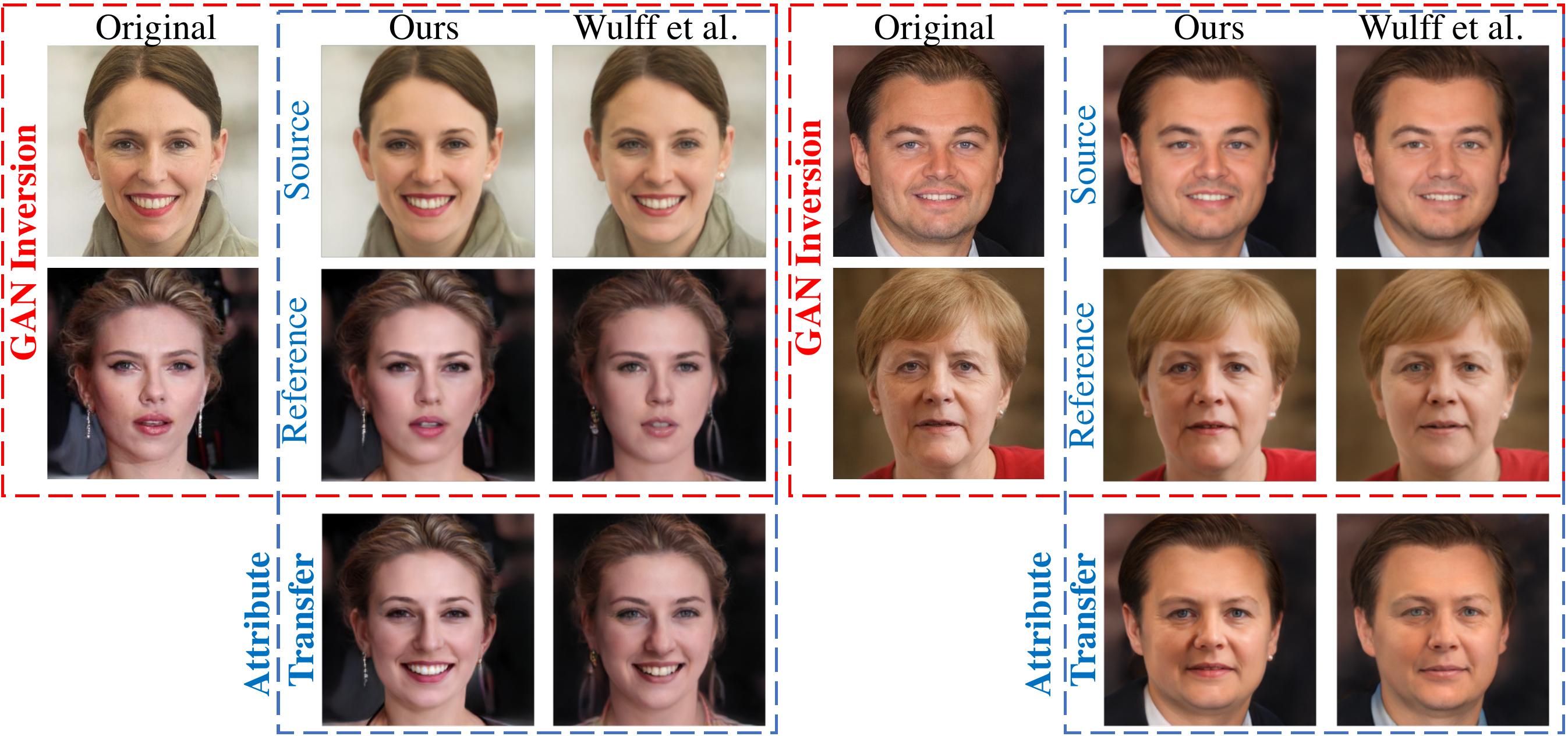}
    \vspace{-0mm}
    \caption{
    \textbf{GAN Inversion.}
    We compare our GAN inversion vs SOTA Wulff et al.~\cite{wulff2020improving}. Our method more faithfully reconstruct the original image while maintaining better editability. Our deepfakes are more natural and captures the facial attributes better.
    }
    \label{fig:compare_inversion}
    \vspace{\figmargin}
\end{figure}

\begin{minipage}[t]{0.46\textwidth}
\centering
\captionof{table}{\textbf{User Preference.}
We conduct user study on different tasks against other methods. 
}
\vspace{-1mm}
\label{tab:user_study}
\begin{tabular}{c|c}
      \multicolumn{2}{c}{\textbf{Attributes Transfer}}\\
      \hline
       vs. Suzuki et al.~\cite{suzuki2018spatially} & vs. EIS~\cite{collins2020editing} \\
       \hline
       70.4\% & 64.0\% \\
       \hline
        \multicolumn{2}{c}{\rule{0pt}{2ex}\textbf{Feature Interpolation}}\\
       \hline
        \multicolumn{2}{c}{vs. Suzuki et al.~\cite{suzuki2018spatially}} \\
        \hline
         \multicolumn{2}{c}{87.6\%}\\
         \hline
       \multicolumn{2}{c}{\rule{0pt}{2ex}\textbf{Single Image Generation}} \\
      \hline
     \multicolumn{2}{c}{vs. SinGAN~\cite{rottshaham2019singan}}\\
    \hline
     \multicolumn{2}{c}{83.3\%} \\
  \end{tabular}
\comment{
  \begin{tabular}{c|c|c|c}
      \multicolumn{2}{c|}{\textbf{Attributes Transfer}} &
      \textbf{Feature Interpolation} &
      \textbf{Single Image Generation} \\
      \hline
     vs. Suzuki et al.~\cite{suzuki2018spatially} & vs. EIS~\cite{collins2020editing} & vs. Suzuki et al.~\cite{suzuki2018spatially} & vs. SinGAN~\cite{rottshaham2019singan}\\
    \hline
     70.4\% & 64.0\% & 87.6\% & 83.3\% \\
  \end{tabular}
  }
\end{minipage}
\hfill
\begin{minipage}[t]{0.46\textwidth}
\centering
\captionof{table}{\textbf{Quantitative Results on Panorama Generation.}
We measure $\infty$-FID to evaluate the visual quality of the generated panorama.
}
\vspace{-1mm}
\centering
\label{tab:fid_table}
\begin{tabular}{c|c}
Method & FID\\
\hline
StyleGAN2~\cite{karras2020analyzing} & 4.5 \\ \\
Method & $\infty$-FID\\
\hline
ALIS~\cite{skorokhodov2021aligning}& 10.5\\
Ours & 15.7 \\ 
Ours + latent smoothing & 12.9\\
\end{tabular}
\end{minipage}

\vspace{\subsecmargin}
\subsection{Improved GAN inversion}
\label{subsec:inv}
\vspace{\subsecmargin}
GAN inversion aims to locate a style code in the $\mathcal{W}+$ space that can synthesize an image similar to the given target image.
In practice, despite being able to reconstruct the target image, the resulting style codes often fall into unstable out-of-domain regions of the space, making it difficult to perform any semantic control over the resulting images. Wulff et al~\cite{wu2020stylespace} discover that under a simple non-linear transformation, the $\mathcal{W}+$ space can be modeled with a Gaussian distribution. Applying a Gaussian prior improves the stability of GAN inversion. However, in our attributes transfer setting, we need to invert both a source and reference image, this formulation struggles to provide satisfactory results.

In a StyleGAN, the $\mathcal{W}$ latent space is mapped to the style coefficients space $\bsigma$ by an affine transformation in the AdaIN module. Recent work has shown better performance in face manipulations~\cite{xu2020generative, collins2020editing} utilizing $\bsigma$ compared to $\mathcal{W}+$. We discover that the $\bsigma$ space without any transformations can also be modeled as a Gaussian distribution. We are then able to impose the same Gaussian prior in this space instead during GAN inversion.

In Fig.~\ref{fig:compare_inversion}, we compare our GAN inversion with Wulff et al. and show significant improvements in the reconstruction and editability of the image. For both GAN inversions, we perform $3000$ descent steps with LPIPS~\cite{zhang2018perceptual} and MSE loss.

\begin{figure}[t]
    \centering
    \includegraphics[width=1\linewidth]{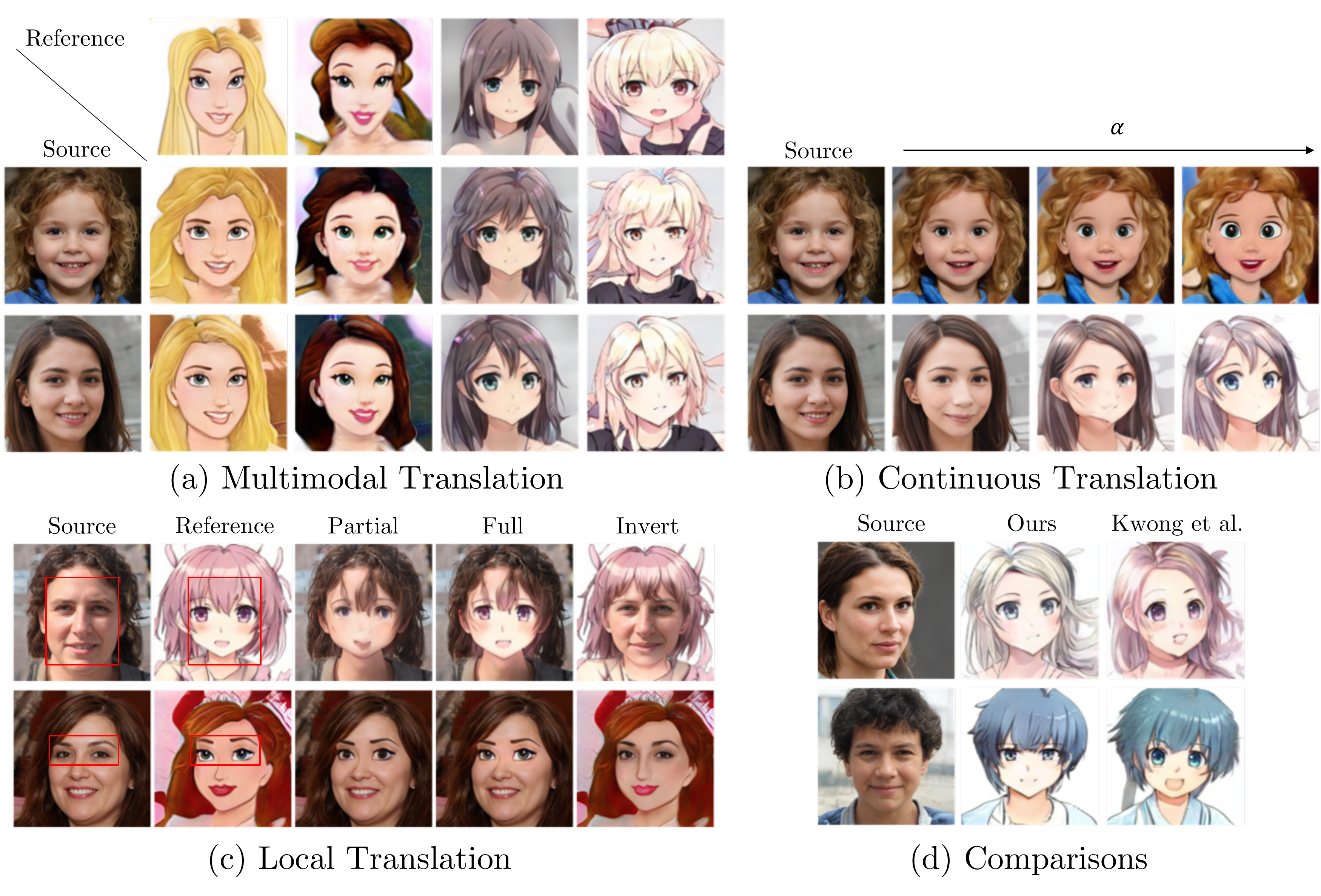}
    \caption{
    \textbf{Image-to-Image Translation.}
    (a) Our method easily transfers to multimodal image translation on multiple datasets. (b) We can control the degree translation. (c) We are also able to perform local translation on a user-prescribed region. (d) Our method preserves semantics better compared to Kwong et al.~\cite{kwong2021unsupervised} (note the facial expressions).}
    \label{fig:compare_i2i}
    \vspace{\figmargin}
\end{figure}

\vspace{\subsecmargin}
\subsection{Controllable I2I translation}
\vspace{\subsecmargin}
Building upon Toonify, Kwong et al.~\cite{kwong2021unsupervised} propose to freeze the fully-connected layers during finetuning phase to better preserve semantics after I2I translation. This preserves StyleGAN's $\mathcal{W}+$ space, which exhibits disentanglement properties \cite{karras2020analyzing, shen2020interpreting, abdal2019image2stylegan}. Following the discussion in \subsecref{inv} that $\bsigma$ space exhibits better disentanglement compared to $\mathcal{W}+$ space, we propose to also freeze the affine transformation layer that produces $\bsigma$. In Fig.~\ref{fig:compare_i2i}(d), we show that this simple change allows us to better preserve the semantics for image translation (note the expressions and shapes of the mouths). 

Following Toonify, we first finetune an FFHQ-pretrained StyleGAN on the target dataset. Both Toonify and Kwong et al. then proceed to perform weight swapping for I2I. While they produce visually pleasing results, they have limited control over the degree of image translation. One interesting observation we make is that feature interpolation also works across the pretrained and finetuned StyleGAN. This allows us to blend real and Disney faces together in numerous ways, achieving different results: 1) We can perform \emph{continuous translation} by using a constant $\alpha$ across all spatial dimensions. The value of $\alpha$ determines the degree of translation. 2) We can perform \emph{localized image translation} by choosing which area to perform feature interpolation. 
3) We can use GAN inversion to perform both face editing and translation on real faces. Using our improved GAN inversion allows more realistic and accurate results.

\begin{figure}[t]
    \centering
    \includegraphics[width=1\linewidth]{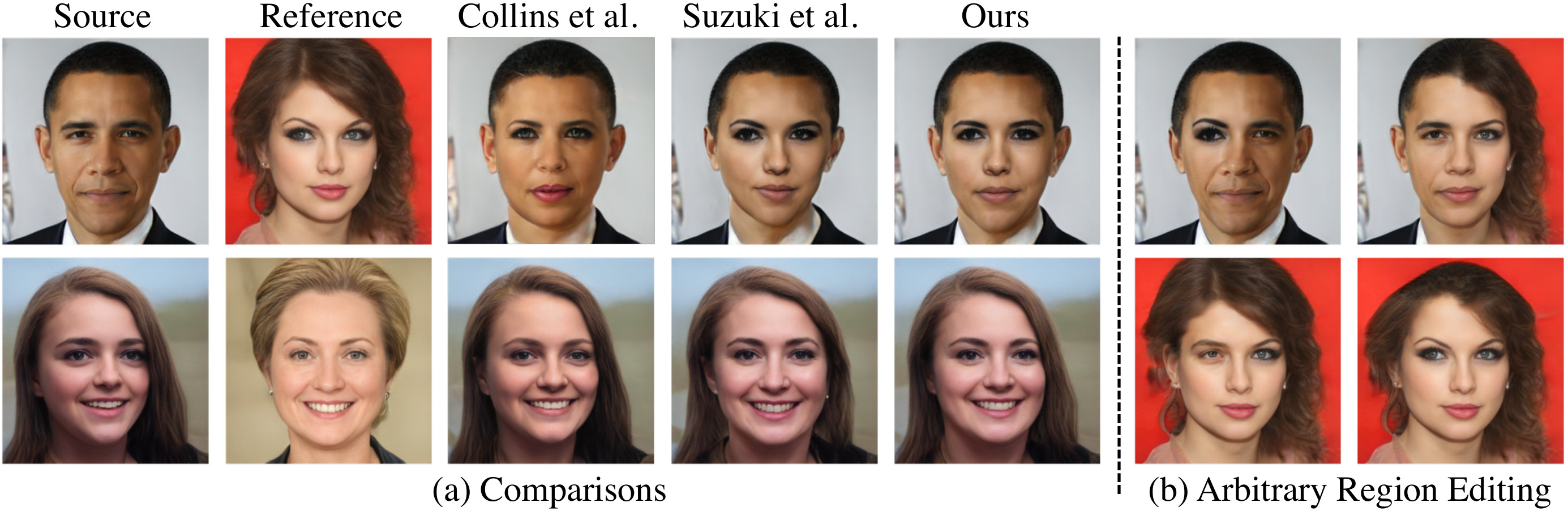}
    \caption{
    \textbf{Attribute Transfer Comparisons.} We compare attribute transfer with other state-of-the-art methods. Collins et al.~\cite{collins2020editing} does not accurately transfers fine-grained attributes, and Suzuki et al.~\cite{suzuki2018spatially} produces unrealistic outputs when the poses are mismatched. Our method is both accurate and realistic. Furthermore, our method is also able to perform transfer in arbitrary regions. We can seamlessly blend two halves of a face, have two distinctly different eyess on each side, etc.
    }
    \vspace{\figmargin}
    \label{fig:compare_deepfake}
\end{figure}

Fig.~\ref{fig:compare_i2i} shows a comprehensive overview of our capabilities in I2I translations. We show that we can perform multimodal translations across different datasets. Reference images provide the overall style of the translated image, while source images provide semantics such as pose, hair style, etc. Sampling different reference images also results in significantly varied styles (drawing style, colors, etc). By controlling $\alpha$ blending parameter, we also show visually pleasing continuous translation results. For example, in the first row of Fig.~\ref{fig:compare_deepfake}(b), we can maintain the texture of a real face while enlarging the eyes. We further show that we can selectively choose which area to translate through feature interpolation. This gives us a large degree of controllability, allowing us to create a face with Disney eyes or even an anime head with a human face. 

\begin{figure}[t]
    \centering
    \includegraphics[width=1\linewidth]{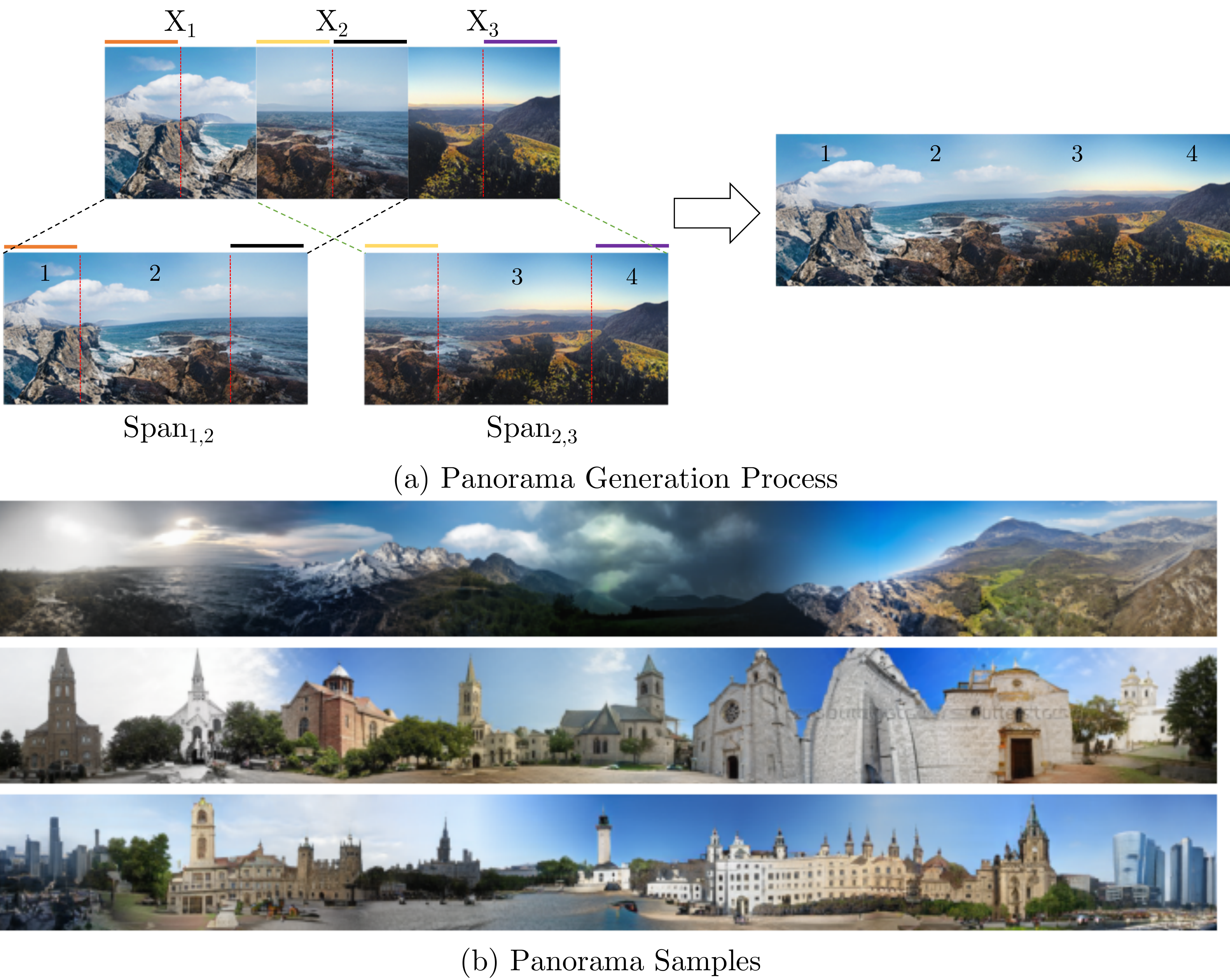}
    \vspace{-0mm}
    \caption{
    \textbf{Panorama Generation.}
    We generate panoramas by knitting spans (blends of two images). We enforce certain constraints to enable flawless knitting. Colored bars represent that the areas are exactly the same, the numbers represent the areas we take to obtain the final panorama. By ensuring the yellow area of $X_2$ is the same as in Span$_{2,3}$, black area of $X_2$ same as in Span$_{1,2}$, we can knit Span$_{1,2}$ and Span$_{2,3}$ perfectly. We can repeat this process to form an arbitrary-lengthed panorama. We show random samples from LHQ, LSUN-Churches, and LSUN-Towers.
    }
    \label{fig:panorama}
    \vspace{\figmargin}
\end{figure}

\vspace{\subsecmargin}
\subsection{Panorama Generation}
\vspace{\subsecmargin}
Using feature interpolation, we can blend two side-by-side images by creating a realistic transition that connects them. We can extend this into infinite panorama generation by continuously blending two images and knitting them together. Under certain blending constraints illustrated in Fig.~\ref{fig:panorama}, we can knit them perfectly. To enforce the constraint that specified areas remain the same, we can choose which areas to blend by a careful choice of $\alpha$ weights. Note that we are not limited to blending only two images at once. The limitation is induced by the GPU memory. 
Depending on the dataset, our panorama method is not limited to horizontal generation and can be extended in any direction.

Even though feature interpolation allows us to blend images that are different, the results are not ideal when the input images are too semantically dissimilar (e.g. side-by-side blending of sea and trees). To overcome this issue, we perform \emph{latent smoothing} -- applying a Gaussian filter across latent codes to smooth neighboring latent codes. It results in more similar neighboring images and as such, have a more natural interpolation between them, leading to more natural results.

In the experiment, for blending images to form a panorama, we perform feature interpolation at every single layer. We choose a blending mask $\alpha$ by linearly scaling it from left to right in the areas constraint by our construction in Fig.~\ref{fig:panorama}. We quantitatively compare our method with ALIS~\cite{skorokhodov2021aligning} using the $\infty$-FID introduced in it. Just by hijacking a pretrained StyleGAN, our method is able to obtain comparable $\infty$-FID with ALIS, which is trained specifically for this task. We also show that performing latent smoothing leads to significant improvement in the score.
\begin{figure}[t]
    \centering
    \includegraphics[width=1\linewidth]{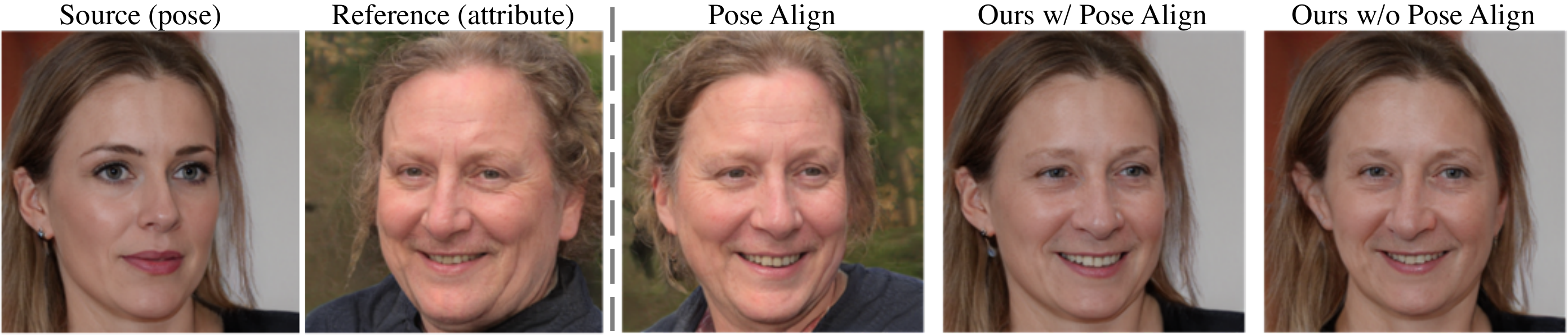}
    \vspace{-0mm}
    \caption{
    \textbf{Attributes Transfer with Pose Align.}
    Naive feature interpolation does not work well when images have very different poses. Our method addresses the problem with a simple pose alignment that allowing us to perform attributes transfers regardless of original poses. 
    }
    \label{fig:pose_align}
    \vspace{\figmargin}
\end{figure}

\vspace{\subsecmargin}
\subsection{Attributes Transfer}
\vspace{\subsecmargin}
While Suzuki et al.~\cite{suzuki2018spatially} show that feature collaging can perform localized feature transfer between two images, the results are highly dependent on pose and orientation. Transferring features from a left-looking face to a right-looking face will cause awkward misalignments. Similarly, naively applying our feature interpolation leads to similar results. EIS allows realistic facial feature transfer that performs well even when faces have different poses. However, EIS does not ensure that irrelevant regions are not affected, e.g., transferring eye features can affect the nose features too. Moreover, EIS only allows transfers for predefined features and not arbitrary user-defined features. Lastly, EIS only allows generating in-distribution images, limiting its ability to generate less common examples such as having one eye with makeup and one without.

In order to allow feature interpolation to work well for arbitrary poses, we perform a pose alignment between source and reference images. There are numerous ways to pose align for StyleGAN images~\cite{shen2020interfacegan, harkonen2020ganspace}. Based on the observation in \cite{karras2020analyzing} that early layers of StyleGAN primarily control pose and structure, we can simply align the first $2048$ dimensions of the $\mathcal{W+}$ style code between the source and reference images. Once pose aligned, we can then apply feature interpolation to transfer chosen features from reference to source. This procedure is shown in Fig.~\ref{fig:pose_align}.

We can further allow arbitrary localized edits by choosing which area to perform feature interpolation. The final pipeline involves a user drawing a bounding box on the source face they wish to change (say eyes + nose). Attributes will then be automatically transferred from a chosen reference face even if their poses are not aligned. We can even generate interesting out-of-distribution examples such as a vertical blending between a male and female face Fig.~\ref{fig:compare_deepfake}(b).

To perform natural attributes transfer with minimal blending artifacts, we perform feature interpolation on layers $i \leq 12$. In Fig.~\ref{fig:compare_deepfake} we qualitatively compare our face attributes transfer method with several other methods. We use the proposed improved GAN inversion method to perform the comparisons on real images. Our results are generally more realistic and better capture the attributes we are interested in. Suzuki et al. produce unnatural images due to the difference in poses between source and reference images, while EIS is less accurate in transferring attributes. We further validated our results through a user study where users choose based on both realism and transfer accuracy, Table~\ref{tab:user_study}. Our method is preferred by the users over both other methods.

\section{Conclusions and Broader Impacts}\label{sec:broader_impacts}
\vspace{\secmargin}
In this work, we show that with only pretrained StyleGAN models along with the proposed spatial operations on the latent space, we can achieve comparable results in various image manipulation tasks that usually require task-specific architectures or training paradigms.
The proposed method is lightweight, efficient, and applicable to any pretrained StyleGAN model.

Our method provides a simple and computationally efficient procedure for general public to perform a variety of image manipulation tasks. However, as a trade-off, this method can also just as easily be applied for disinformation. For example, attributes transfer can be used to make DeepFakes which can be used maliciously. Also, as our method relies on a pretrained StyleGAN, it is also limited by the capacity of it. There may be issues of diversity where minorities are not well represented in the dataset. As such, our method might not be able to perform manipulations well on faces of minorities. A well balanced dataset that properly represents the minorities is pertinent to a fair model. More research and insight into mode dropping in GANs are also necessary. 
\clearpage

\bibliographystyle{unsrt}
\bibliography{bibliography}
\appendix

\newpage
\section{Appendix}\label{sec:appendix}
\vspace{\secmargin}

\subsection{$\alpha$ blending}
\begin{figure}[ht]
    \centering
    \includegraphics[width=1\linewidth]{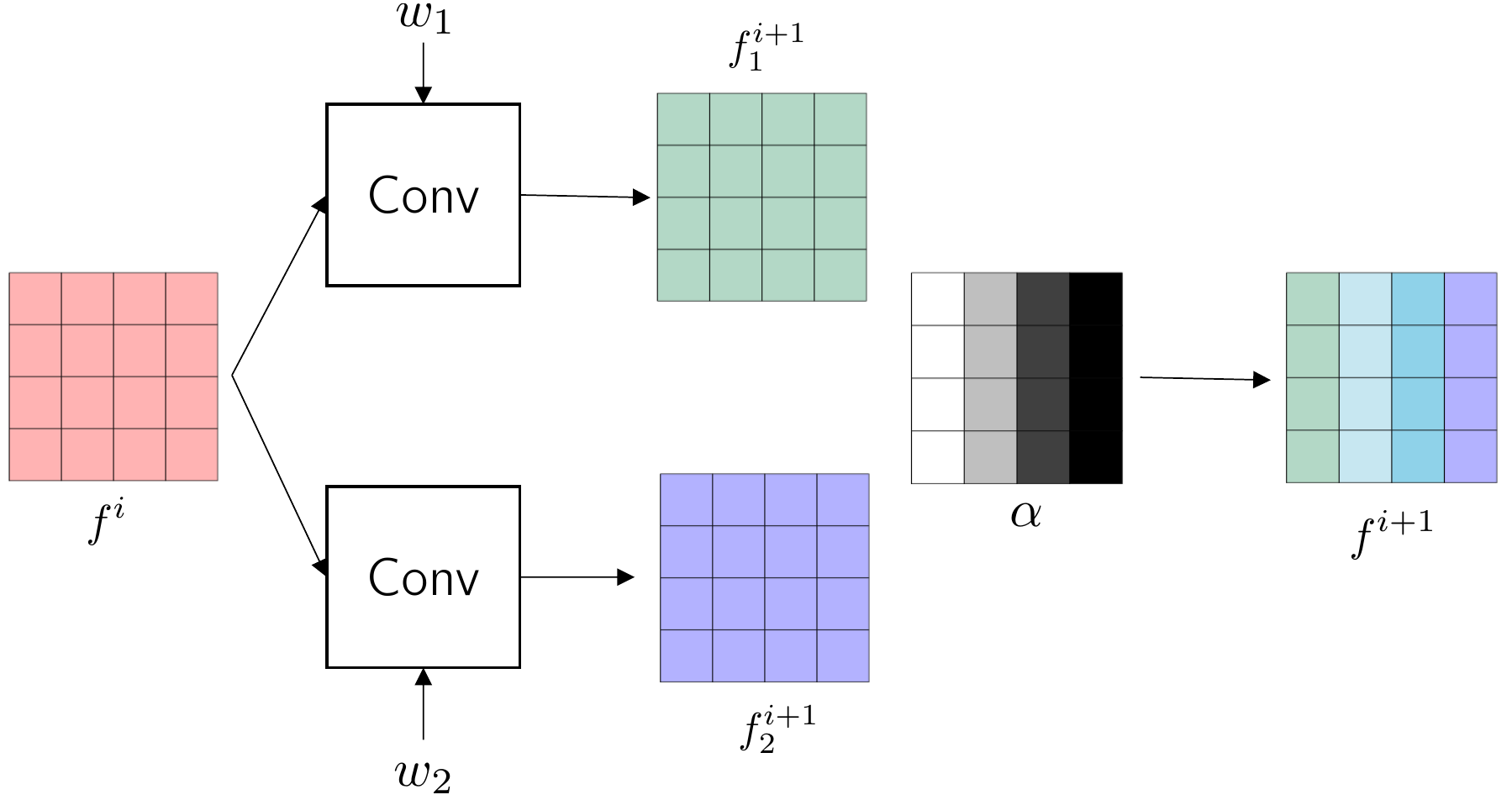}
    \caption{$\alpha$ \textbf{blending.}
    We inject 2 different styles to get 2 intermediate features $f_1^{i+1}$ and $f_2^{i+1}$, which we blend using a spatially varying $\alpha$ mask. The final output is then passed on to the next convolution layers where the same process is repeated.
    }
    \label{fig:supp_procedure}
\end{figure}

\begin{figure}[ht]
    \centering
    \includegraphics[width=1\linewidth]{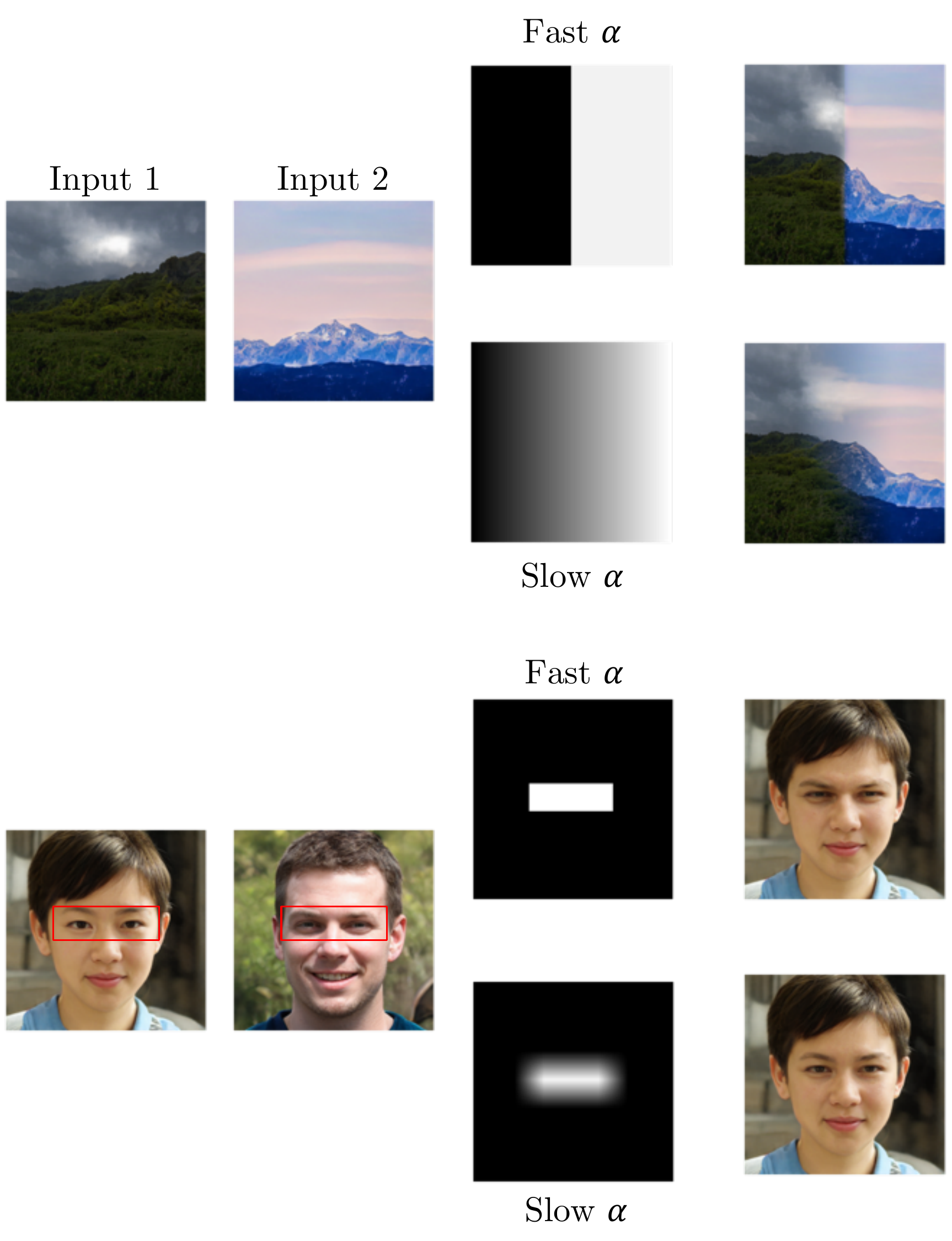}
    \caption{$\alpha$\textbf{scaling.}
    Different $\alpha$ scaling gives different blend results. Fast scaling $\alpha$ preserves features better. This is harmful if the two images are very different as the transition will be abrupt as seen in the landscape example. It is however useful for accurate facial attributes transfer. Slow scaling $\alpha$ gives a slow smooth blend which is helpful for landscapes, but fails to accurately preserves facial features.    }
    \label{fig:supp_alpha_scale}
\end{figure}

We blend images with an $\alpha$ mask. 
We can control different speed of scaling from 0 to 1 to obtain different $\alpha$ masks for feature blending.
In \figref{supp_procedure}, we illustrate the concept of alpha blending. In \figref{supp_alpha_scale}, we apply different alpha masks to different tasks. 
For landscape images where contents are usually structurally different, slower $\alpha$ allows smoother transition. 
On the other hand, for face editing, faster $\alpha$ is usually beneficial as we want to accurately reproduce the fine-grained features from the reference without it being affected by the transitions.

\subsection{Latent smoothing}
\begin{figure}[hb]
    \centering
    \includegraphics[width=1\linewidth]{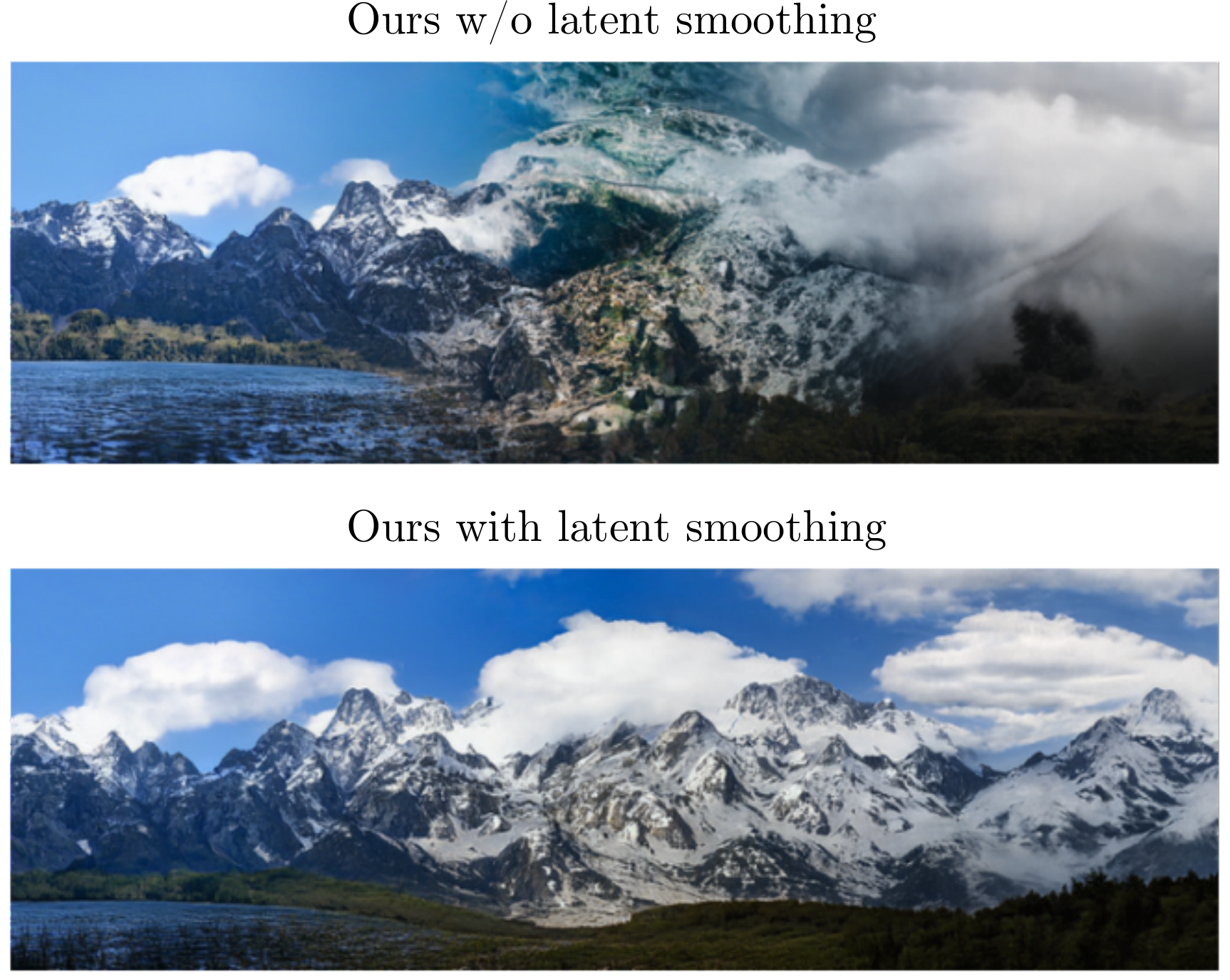}
    \caption{\textbf{Latent smoothing.}
    We compare our feature interpolation with and without latent smoothing. When blending 2 very different images, the resulting blend can be unrealistic (left). Latent smoothing causes neighboring latent codes (and consequently neighboring images) to be closer, giving a more realistic image blending. 
    }
    \vspace{\figmargin}
    \label{fig:supp_latent_smooth}
\end{figure}

In addition to feature interpolation, we adopt latent smoothing to handle cases that input images are too semantically dissimilar.
We apply a Gaussian filter across latent codes.
As shown in \figref{supp_latent_smooth}, latent smoothing can greatly alleviate the artifacts.

\subsection{More Samples}
\begin{figure}[t]
    \centering
    \includegraphics[width=1\linewidth]{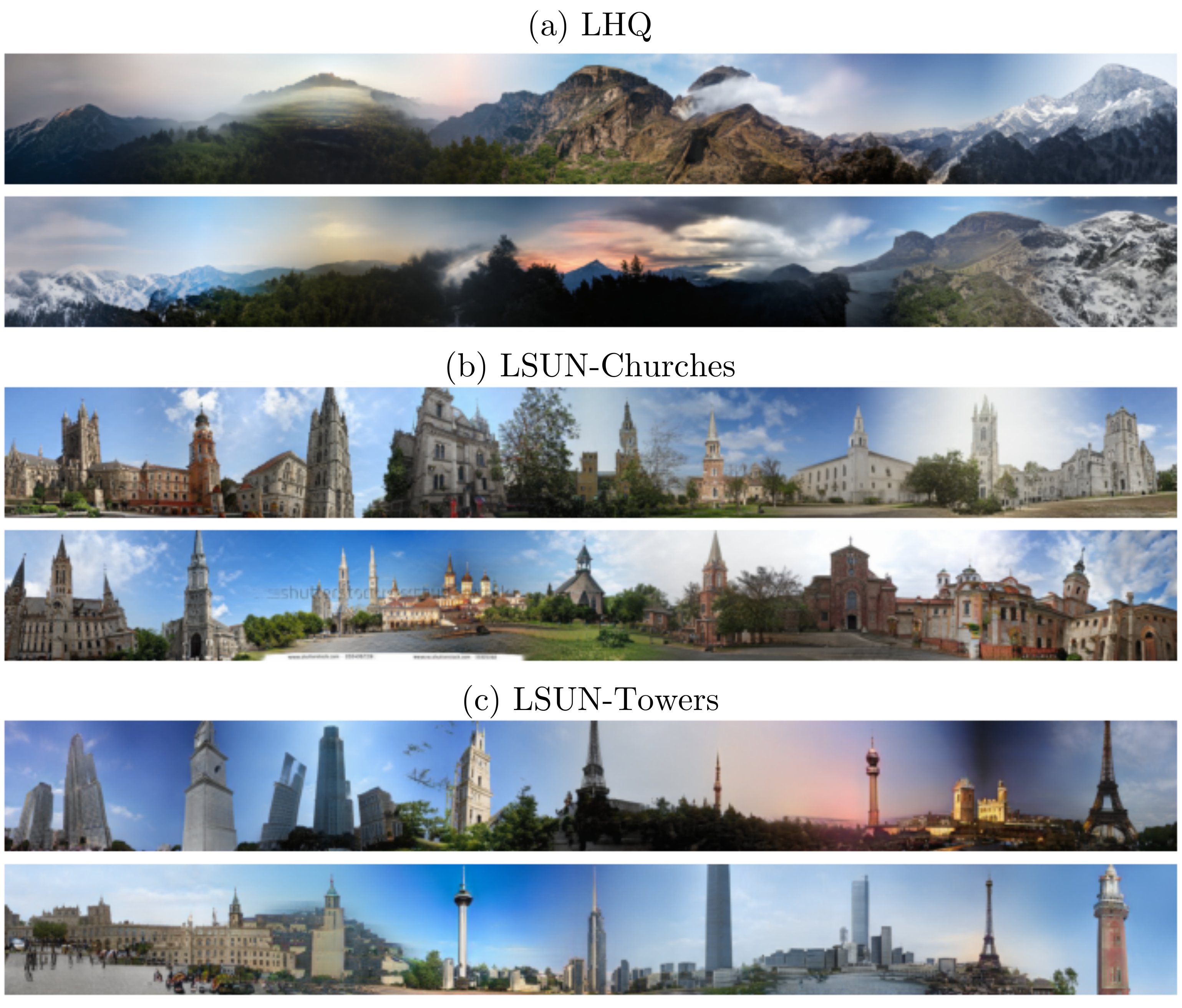}
    \caption{\textbf{More Samples of Panorama generation.} 
    }
    \vspace{\figmargin}
    \label{fig:supp_panorama}
\end{figure}

\begin{figure}[t]
    \centering
    \includegraphics[width=1\linewidth]{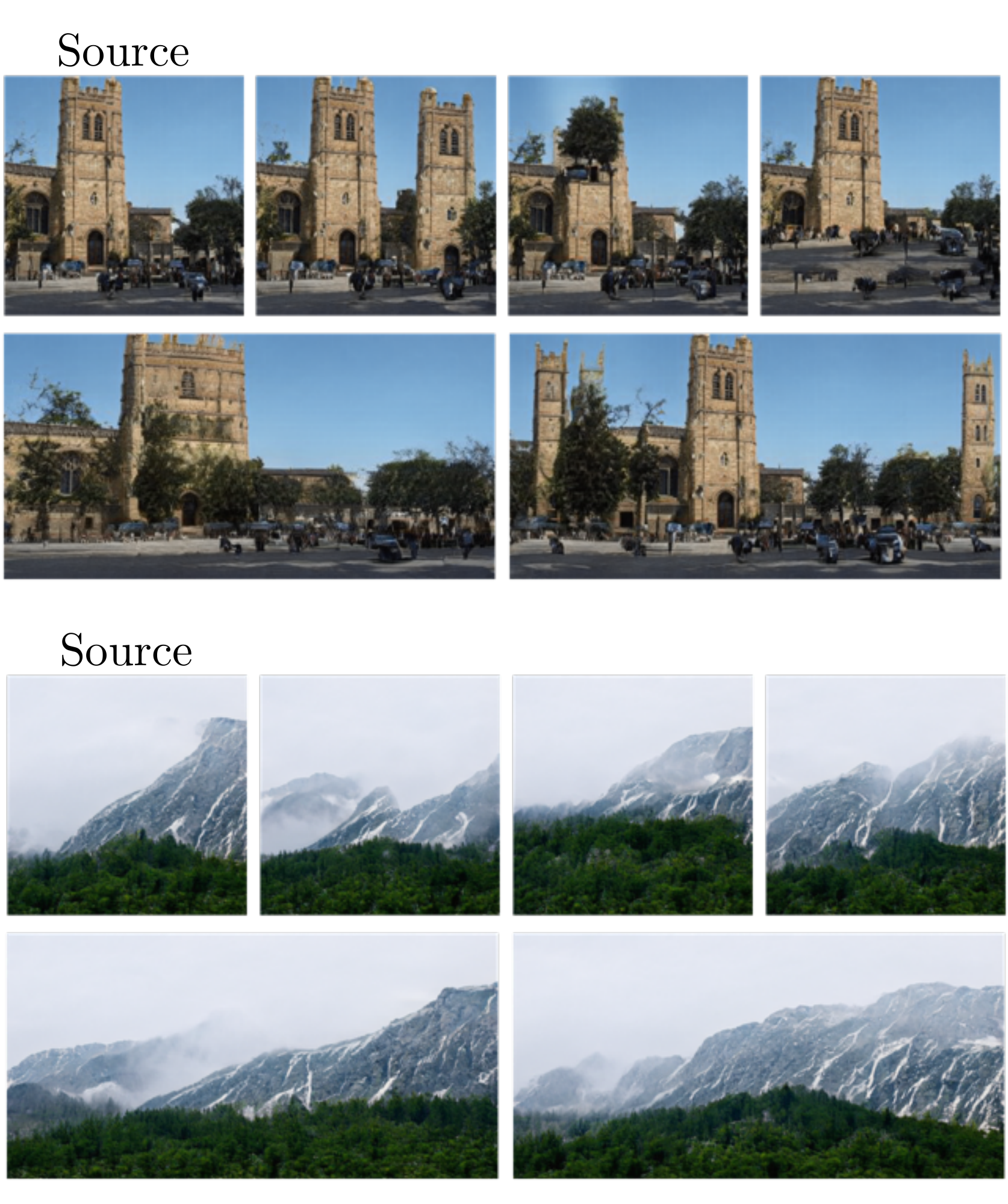}
    \caption{\textbf{More Samples of Generation from a single image.}
    }
    \vspace{\figmargin}
    \label{fig:supp_single_im}
\end{figure}

\begin{figure}[t]
    \centering
    \includegraphics[width=1\linewidth]{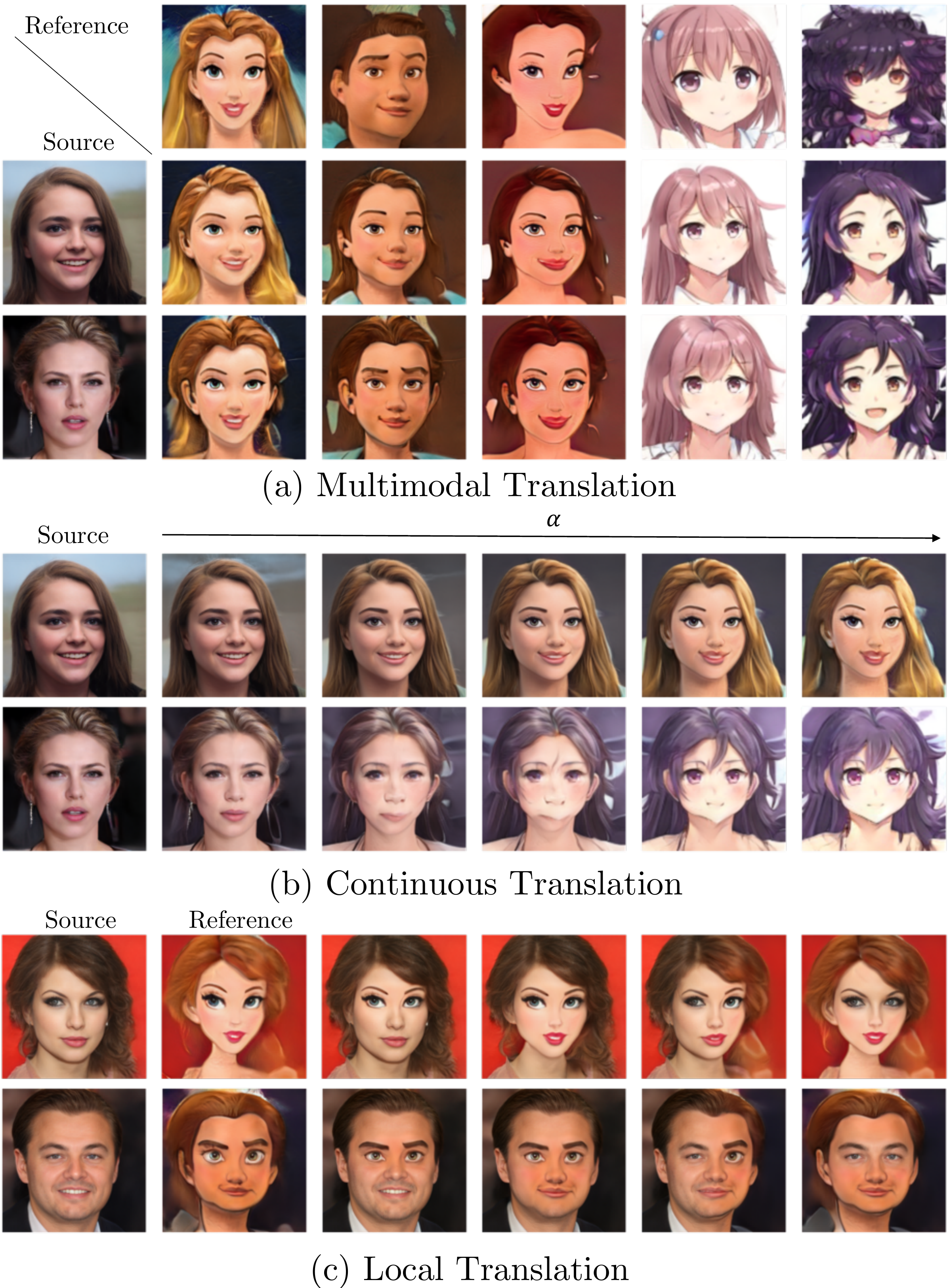}
    \caption{
    \textbf{More Samples of Image-to-Image translation.}
    }
    \vspace{\figmargin}
    \label{fig:supp_face_translate}
\end{figure}

We present more samples on parorama generation, generation from a single image, and image-to-image translation in \figref{supp_panorama}, \figref{supp_single_im}, and \figref{supp_face_translate}, respectively.

\end{document}